\documentclass[11pt,a4paper]{article}
\usepackage[hyperref]{emnlp2020}
\usepackage{times}
\usepackage{latexsym}

\usepackage{multirow}
\usepackage{amsmath}
\usepackage{graphicx}
\usepackage{amsthm}
\usepackage{latexsym}
\usepackage{amsfonts}
\usepackage{times}
\usepackage{floatrow}
\usepackage{mathtools}
\usepackage{algorithm2e}
\usepackage{algorithmic}
\usepackage{bbm}
\RestyleAlgo{algoruled}
\usepackage{pacioli}
\usepackage{fetamont}
\usepackage{color}
\usepackage{todonotes}
\usepackage[normalem]{ulem}
\usepackage{floatrow}

\usepackage[title]{appendix}

\usepackage{caption} \usepackage{subcaption}

\SetKwInput{KwInput}{Input}
\SetKwInput{AKwInput}{Extra Input}
\SetKwInput{KwOutput}{Output}
\SetKwBlock{Repeat}{Repeat k time steps}{}
\SetKwBlock{Rep}{Repeat (k+d-1) time steps}{}
\SetKwInput{KwInitial}{Initialization}
\SetKwInput{AKwInitial}{Extra Initialization}
\newcommand{\X}{\mathbb{S}}
\newcommand{\T}{\mathbb{T}}
\newcommand{\A}{$\mathbb{A}$}
\newcommand{\B}{$\mathbb{B}$}
\newcommand{\C}{$\mathbb{C}$}

\usepackage{mflogo}
\newcommand{\BAE}[1]{{\texttt{BAE#1}}}
\newcommand{\BAEfont}[1]{{\texttt{#1}}}

\newcommand{\blue}[1]{\textcolor{blue}{#1}}
\newcommand{\red}[1]{\textcolor{red}{#1}}

\newcommand{\lll}{\mathbb{L}}

\usepackage{amsmath}
\usepackage{mathtools}

\usepackage{booktabs}
\usepackage{multirow}
\usepackage{microtype}

\aclfinalcopy 
\def\aclpaperid{371} 

\newcommand{\myparagraph}[1]{\noindent\textbf{#1. } }

\title{\BAE{}: BERT-based Adversarial Examples for Text Classification}

\author{Siddhant Garg\thanks{\ \ \  Equal contribution by authors} \ \thanks{\ \ \ Work completed as a graduate student at UW-Madison}\\
Amazon Alexa AI Search\\
Manhattan Beach, CA, USA\\
 \texttt{sidgarg@amazon.com} \\
\And Goutham Ramakrishnan\footnotemark[1] \ \footnotemark[2] \\
 Health at Scale Corporation \\
 San Jose, CA, USA\\
 \texttt{gouthamr@cs.wisc.edu} \\
}

\begin{document}
\maketitle
\begin{abstract}
Modern text classification models are susceptible to adversarial examples, perturbed versions of the original text indiscernible by humans which get misclassified by the model.
Recent works in NLP use rule-based synonym replacement strategies to generate adversarial examples.
These strategies can lead to out-of-context and unnaturally complex token replacements, which are easily identifiable by humans.
We present \BAE{}, a black box attack for generating adversarial examples using contextual perturbations from a BERT masked language model.
\BAE{} replaces and inserts tokens in the original text by masking a portion of the text and leveraging the BERT-MLM to generate alternatives for the masked tokens. 
Through automatic and human evaluations, we show that \BAE{} performs a stronger attack, in addition to generating adversarial examples with improved grammaticality and semantic coherence as compared to prior work.
\end{abstract}

\section{Introduction}
Recent studies have exposed the vulnerability of ML models to adversarial attacks, small input perturbations which lead to misclassification by the model. Adversarial example generation in NLP~\cite{DBLP:journals/corr/abs-1901-06796} is more challenging than in commonly studied computer vision tasks~\cite{42503,45818,Papernot:2017:PBA:3052973.3053009} because of (i) the discrete nature of the input space and (ii) the need to ensure semantic coherence with the original text. 
A major bottleneck in applying gradient based~\cite{43405} or generator model~\cite{zhao2018generating} based approaches to generate adversarial examples in NLP is the backward propagation of the perturbations from the continuous embedding space to the discrete token space.

Initial works for attacking text models relied on introducing errors at the character level~\cite{ebrahimi-etal-2018-hotflip,DBLP:journals/corr/abs-1801-04354} or adding and deleting words~\cite{DBLP:journals/corr/LiMJ16a,DBLP:journals/corr/LiangLSBLS17,DBLP:journals/corr/abs-1804-07781} for creating adversarial examples. 
These techniques often result in unnatural looking adversarial examples which lack grammatical correctness, thereby being easily identifiable by humans.

\begin{figure}[t]
    \centering
    \includegraphics[width=\linewidth]{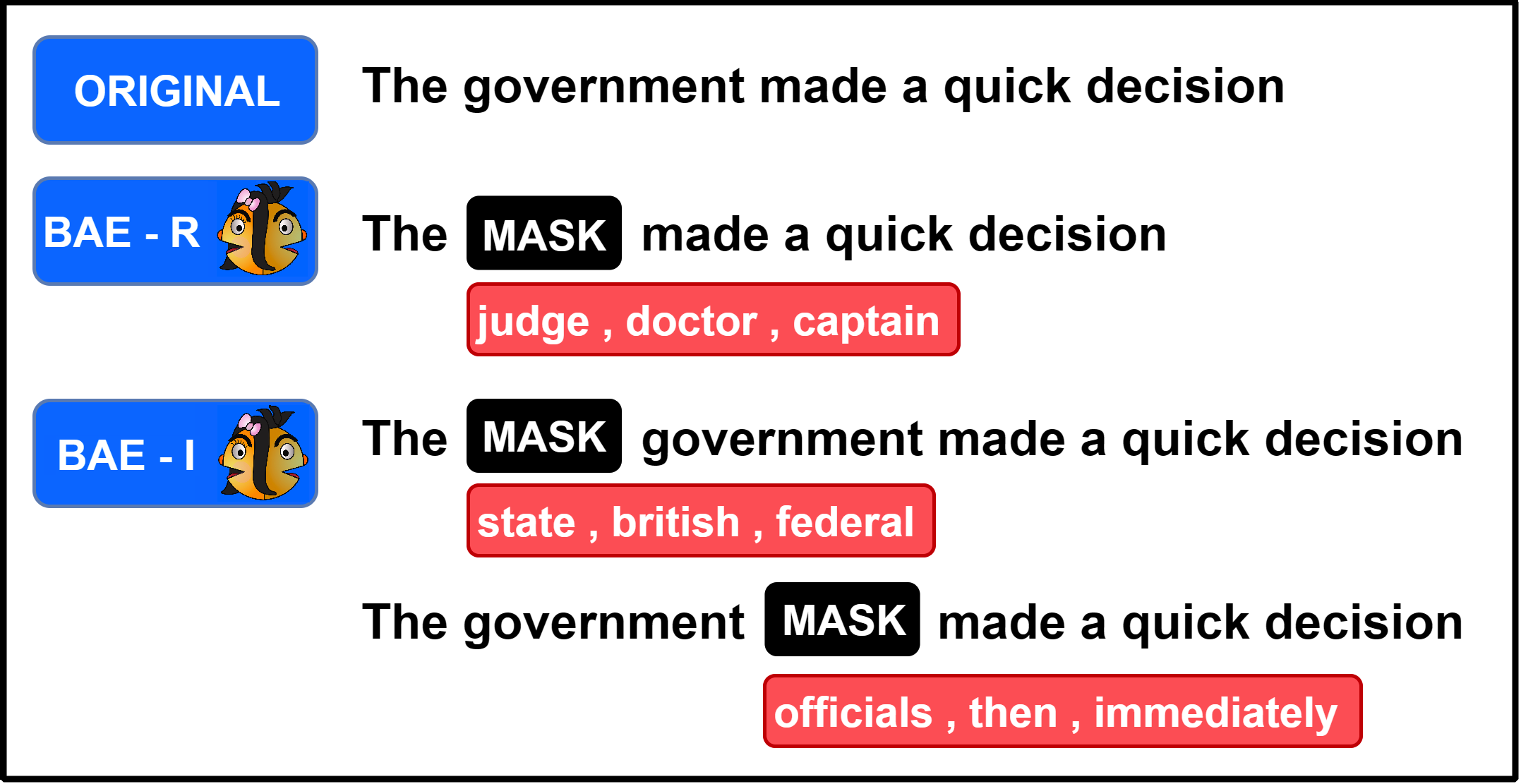}
    \vspace{-1.5em}
    \caption{We use BERT-MLM to predict masked tokens in the text for generating adversarial examples.
    The MASK token replaces a word (\BAE{-R} attack) or is inserted to the left/right of the word (\BAE{-I}).}
    \label{fig:BAE_Attack}
\end{figure}

Rule-based synonym replacement strategies ~\cite{alzantot-etal-2018-generating,ren-etal-2019-generating} have recently lead to \textit{more} natural looking adversarial examples. \citet{jin2019bert} combine both these works by proposing TextFooler, a strong black-box attack baseline for text classification models.
However, the adversarial examples generated by TextFooler solely account for the token level similarity via word embeddings, and not the overall sentence semantics. 
This can lead to \emph{out-of-context} and \emph{unnaturally complex} replacements (see Table~\ref{tab:examples}), which are easily human-identifiable.
Consider a simple example: ``The restaurant service was \emph{poor}". 
Token level synonym replacement of `poor' may lead to an inappropriate choice such as `broke', while a context-aware choice such as `terrible' leads to better retention of semantics and grammaticality. 

Therefore, a token replacement strategy contingent on retaining sentence semantics using a powerful language model~\cite{devlin2018bert,radford2019language} can alleviate the errors made by existing techniques for homonyms (tokens having multiple meanings).
In this paper, we present \BAE{} (BERT-based Adversarial Examples), a novel technique using the BERT masked language model (MLM) for word replacements to better fit the overall context of the English language. 
In addition to replacing words, we also propose inserting new tokens in the sentence to improve the attack strength of \BAE{}. 
These perturbations in the input sentence are achieved by masking a part of the input and using a LM to fill in the mask (See Figure~\ref{fig:BAE_Attack}).

Our \BAE{} attack beats the previous baselines by a large margin on empirical evaluation over multiple datasets and models. 
We show that, surprisingly, just a few replace/insert operations can reduce the accuracy of even a powerful BERT classifier by over $80\%$ on some datasets.
Moreover, our human evaluation reveals the improved grammaticality of the adversarial examples generated by \BAE{} over the baseline TextFooler, which can be attributed to the BERT-MLM.
To the best of our knowledge, we are the first to use a LM for generating adversarial examples. 
We summarize our contributions as:
\begin{itemize}
    \item \vspace{-0.3em} We propose \BAE{}, an adversarial example generation technique using the BERT-MLM. 
    \item \vspace{-0.3em} We introduce 4 \BAE{} attack modes by replacing and inserting tokens, all of which are almost always stronger than previous baselines on 7 text classification datasets.
    \item \vspace{-0.3em} Through human evaluation, we show that \BAE{} yields adversarial examples with improved grammaticality and semantic coherence.
    \end{itemize}
 
\section{Methodology}
\vspace{-0.4em}
\myparagraph{Problem Definition}
We are given a dataset $(S,Y) = \{(\X_1,y_1),\dots (\X_m,y_m)\}$ and a trained classification model $C:\X \rightarrow Y$.
We assume the soft-label black-box setting where the attacker can only query the classifier for output probabilities on a given input, and does not have access to the model parameters, gradients or training data. 
For an input pair $(\X{=}[t_1, \dots , t_n],y)$, we want to generate an adversarial example $\X_{adv}$ such that $C(\X_{adv}){\neq}y$. Additionally we would like $\X_{adv}$ to be grammatically correct and semantically similar to $\X$. 

\myparagraph{\BAE{}} For generating an adversarial example $\X_{adv}$, we introduce 2 types of token-level perturbations: (i) Replace a token $t \in \X$ with another and (ii) Insert a new token $t'$ in $\X$.
Some tokens in the input contribute more towards the final prediction by $C$ than others.
Replacing these tokens or inserting a new token adjacent to them can thus have a stronger effect on altering the classifier prediction. 
This intuition stems from the fact that the replaced/inserted tokens changes the local context around the original token.
We estimate token importance $I_i$ of each $t_i \in \X$, by deleting $t_i$ from $\X$ and computing the decrease in probability of predicting the correct label $y$, similar to \citet{jin2019bert,ren-etal-2019-generating}.

The Replace (R) and Insert (I) operations are performed on a token $t$ by masking it and inserting a mask token adjacent to it respectively. The pre-trained BERT-MLM is used to predict the mask tokens (See Figure~\ref{fig:BAE_Attack}) in line with recent work~\cite{shi2020robustness} which uses this to analyse robustness of paraphrase identification models to modifying shared words. 
BERT-MLM is a powerful LM trained on a large training corpus ($\sim$ 2 billion words), and hence the predicted mask tokens fit well into the grammar and context of the text.  





\setlength{\dbltextfloatsep}{3pt}
\renewcommand{\AlCapSty}[1]{\normalfont\small{\textbf{#1}}\unskip}
\begin{algorithm}[t!]
\small
\caption{\small{ \BAE{-R} \ Pseudocode}}
\label{alg1}
\SetAlgoLined
\KwInput{Sentence $\X = [t_1, \dots, t_n]$, ground truth label $y$, classifier model C }
\KwOutput{Adversarial Example $\X_{adv}$}
\KwInitial{$\X_{adv} \leftarrow \X$}

Compute token importance $I_{i} \ \forall \ t_i \in 
\X$

 \For{$i$ \ in descending order of $I_i$ }{
 $\X_M \leftarrow \X_{adv[1:i-1]}[M]\X_{adv[i+1:n]} $\\
 Predict top-K tokens $\T$ for mask M $\in \X_M$ \\
 $\T \leftarrow$ {\sc FILTER}$(\T)$ \\
 $\lll = \{\}$ \tcp{python-style dict}
  \For{$t \in \T$}{
    $\lll[t] = \X_{adv[1:i-1]}[t]\X_{adv[i+1:n]}$
  }

  \textbf{if} $\exists \ t \in \T \ $ s.t $ \ C(\lll[t]) \neq y$ \textbf{then} \\
  \quad \textbf{Return:} \ $\X_{adv} \leftarrow \lll[t']$ where
   $C(\lll[t']) \neq y$,
   \\ \qquad \qquad \ \ $\lll[t']$ has 
   maximum similarity with $\X$\\
  \textbf{else} \\
  \quad $\X_{adv} \leftarrow \lll[t']$ where $\lll[t']$ causes maximum \\ \quad reduction in probability of $y$ in $C(\lll[t'])$ \\ 
  \textbf{end if}
 }
 \textbf{Return:} $\X_{adv} \leftarrow None$ 
\end{algorithm}

\begin{table*}[t]

\begin{center}
\resizebox{0.72\linewidth}{!}{
\begin{tabular}{clllll}
\toprule
\multirow{2}{*}{\textbf{Model}}    & \multicolumn{1}{c}{\multirow{2}{*}{\textbf{Adversarial}}} & \multicolumn{4}{c}{\textbf{Datasets}}                                                                                                         \\ \cmidrule(l){3-6} 
                                   & \multicolumn{1}{c}{\textbf{Attack}}                                             & \multicolumn{1}{c}{\textbf{Amazon}} & \multicolumn{1}{c}{\textbf{Yelp}} & \multicolumn{1}{c}{\textbf{IMDB}} & \multicolumn{1}{c}{\textbf{MR}} \\ \midrule
\multirow{6}{*}{\textbf{wordLSTM}} & Original                                                         & 88.0                                & 85.0                              & 82.0                              & 81.16                           \\
                                   & \small{TextFooler}                                                       & 31.0 \small{(0.747)}                         & 28.0 \small{(0.829)}                       & 20.0 \small{(0.828)}                       & 25.49 \small{(0.906)}                    \\
                                   & \BAE{-R}                                                            & 21.0 \small{(0.827)}                         & 20.0 \small{(0.885)}                       & 22.0 \small{(0.852)}                       & 24.17 \small{(0.914)}                    \\
                                   & \BAE{-I}                                                            & 17.0 \small{(0.924)}                         & 22.0 \small{(0.928)}                       & 23.0 \small{(0.933)}                       & 19.11 \small{(0.966)}                    \\
                                   & \BAE{-R/I}                                                          & 16.0 \small{(0.902)}                         & 19.0 \small{(0.924)}                       & 8.0 \small{(0.896)}                        & 15.08 \small{(0.949)}                    \\
                                   & \BAE{-R+I}                                                          & \textbf{4.0 \small{(0.848)}}                 & \textbf{9.0 \small{(0.902)}}               & \textbf{5.0 \small{(0.871)}}               & \textbf{7.50 \small{(0.935)}}            \\ \midrule
\multirow{6}{*}{\textbf{wordCNN}}  & Original                                                         & 82.0                                & 85.0                              & 81.0                              & 76.66                           \\
                                   & \small{TextFooler}                                                       & 42.0 \small{(0.776)}                         & 36.0 \small{(0.827)}                       & 31.0 \small{(0.854)}                       & 21.18 \small{(0.910)}                    \\
                                   & \BAE{-R}                                                            & 16.0 \small{(0.821)}                         & 23.0 \small{(0.846)}                       & 23.0 \small{(0.856)}                       & 20.81 \small{(0.920)}                    \\
                                   & \BAE{-I}                                                            & 18.0 \small{(0.934)}                         & 26.0 \small{(0.941)}                       & 29.0 \small{(0.924)}                       & 19.49 \small{(0.971)}                    \\
                                   & \BAE{-R/I}                                                          & 13.0 \small{(0.904)}                         & 17.0 \small{(0.916)}                       & 20.0 \small{(0.892)}                       & 15.56 \small{(0.956)}                    \\
                                   & \BAE{-R+I}                                                          & \textbf{2.0 \small{(0.859)}}                 & \textbf{9.0 \small{(0.891)}}               & \textbf{14.0 \small{(0.861)}}              & \textbf{7.87 \small{(0.938)}}            \\ \midrule
\multirow{6}{*}{\textbf{BERT}}     & Original                                                         & 96.0                                & 95.0                              & 85.0                              & 85.28                           \\
                                   & \small{TextFooler}                                                       & 30.0 \small{(0.787)}                         & 27.0 \small{(0.833)}                       & 32.0 \small{(0.877)}                       & 30.74 \small{(0.902)}                    \\
                                   & \BAE{-R}                                                            & 36.0 \small{(0.772)}                         & 31.0 \small{(0.856)}                       & 46.0 \small{(0.835)}                       & 44.05 \small{(0.871)}                    \\
                                   & \BAE{-I}                                                            & 20.0 \small{(0.922)}                         & 25.0 \small{(0.936)}                       & 31.0 \small{(0.929)}                       & 32.05 \small{(0.958)}                    \\
                                   & \BAE{-R/I}                                                          & \textbf{11.0 \small{(0.899)}}                         & 16.0 \small{(0.916)}                       & 22.0 \small{(0.909)}                       & 20.34 \small{(0.941)}                    \\
                                   & \BAE{-R+I}                                                          & 14.0 \small{(0.830)}               & \textbf{12.0 \small{(0.871)}}              & \textbf{16.0 \small{(0.856)}}              & \textbf{19.21 \small{(0.917)}}           \\ \bottomrule
\end{tabular}
}
\end{center}
\vspace{-1em}
\caption{
Automatic evaluation of adversarial attacks on 4 Sentiment Classification tasks. We report the test set accuracy. The average semantic similarity, between the original and adversarial examples, obtained from USE are reported in parentheses. Best performance, in terms of maximum drop in test accuracy, is highlighted in \textbf{boldface}.}
\label{tab:results1}
\end{table*}

The BERT-MLM, however, does not guarantee semantic coherence to the original text as demonstrated by the following simple example. 
Consider the sentence: `the food was good'. 
For replacing the token `good', BERT-MLM may predict the token `bad', which fits well into the grammar and context of the sentence, but changes the original sentiment of the sentence. 
To achieve a high semantic similarity with the original text on introducing perturbations, we filter the set of top K tokens (K is a pre-defined constant) predicted by BERT-MLM for the masked token, using a Universal Sentence Encoder (USE) based sentence similarity scorer~\cite{DBLP:journals/corr/abs-1803-11175}. 
For the R operation, we additionally filter out predicted tokens that do not form the same part of speech (POS) as the original token.

If multiple tokens can cause $C$ to misclassify $\X$ when they replace the mask, we choose the token which makes $\X_{adv}$ most similar to the original $\X$ based on the USE score.
If no token causes misclassification, then we choose the one that decreases the prediction probability $P(C(\X_{adv}){=}y)$ the most. 
We apply these token perturbations iteratively in decreasing order of token importance, until either $C(\X_{adv}){\neq}y$ (successful attack) or all the tokens of $\X$ have been perturbed (failed attack).

We present 4 attack modes for \BAE{} based on the R and I operations, where for each token $t$ in $\X$:
\begin{itemize}
    \item \vspace{-0.2em} \textbf{\BAE{-R}:} Replace token $t$ (See Algorithm~\ref{alg1})
    \item \vspace{-0.5em} \textbf{\BAE{-I}:} Insert a token to the left or right of $t$
    \item \vspace{-0.5em} \textbf{\BAE{-R/I}:} Either replace token $t$ or insert a token to the left or right of $t$
    \item \vspace{-0.5em} \textbf{\BAE{-R+I}:} First replace token $t$, then insert a token to the left or right of $t$
\end{itemize}

\noindent Generating adversarial examples through masked language models has also been recently explored by \citet{li2020contextualized} since our original submission.

\section{Experiments} 
\label{sec:experiment}

\myparagraph{Datasets and Models}
We evaluate \BAE{} on different text classification tasks.
Amazon, Yelp, IMDB are sentiment classification datasets used in recent works~\cite{sarma2018domain} and MR~\cite{Pang+Lee:05a} contains movie reviews based on sentiment polarity. MPQA~\cite{Wiebe2005} is a dataset for opinion polarity detection, Subj~\cite{Pang+Lee:04a} for classifying a sentence as subjective or objective and TREC~\cite{Li:2002:LQC:1072228.1072378} for question type classification.

We use 3 popular text classification models: word-LSTM~\cite{Hochreiter:1997:LSM:1246443.1246450}, word-CNN~\cite{kim2014convolutional} and a fine-tuned BERT~\cite{devlin2018bert} base-uncased classifier. 
We train models on the training data and perform the adversarial attack on the test data. 
For complete model details, refer to Appendix~\ref{app:exp_details}.

As a baseline, we consider TextFooler~\cite{jin2019bert} which performs synonym replacement using
a fixed word embedding space~\cite{mrksic:2016:naacl}. We only consider the top K=50 synonyms
from the BERT-MLM predictions and set a threshold of
0.8 for the cosine similarity between USE based
embeddings of the adversarial and input text.

\newfloatcommand{capbtabbox}{table}[][1.23\linewidth]

\begin{figure*}
\begin{floatrow}
\capbtabbox{%
\resizebox{\linewidth}{!}{    
  \begin{tabular}{cllll}
    \toprule
    \multirow{2}{*}{\textbf{Model}}    & \multicolumn{1}{c}{\multirow{2}{*}{\textbf{Adversarial}}} & \multicolumn{3}{c}{\textbf{Datasets}}                                                                     \\ \cmidrule(l){3-5} 
                                      & \multicolumn{1}{c}{\textbf{Attack}}                                             & \multicolumn{1}{c}{\textbf{MPQA}} & \multicolumn{1}{c}{\textbf{Subj}} & \multicolumn{1}{c}{\textbf{TREC}} \\ \midrule
    \multirow{6}{*}{\textbf{wordLSTM}} & Original                                                         & 89.43                             & 91.9                              & 90.2                              \\
                                      & \small{TextFooler}                                                       & 48.49 \small{(0.745)}                      & 58.5 \small{(0.882)}                       & 42.4 \small{(0.834)}                       \\
                                      & \BAE{-R}                                                            & 45.66 \small{(0.748)}                      & 50.2 \small{(0.899)}                       & 32.4 \small{(0.870)}                       \\
                                      & \BAE{-I}                                                            & 40.94 \small{(0.871)}                      & 49.8 \small{(0.958)}                       & 18.0 \small{(0.964)}                       \\
                                      & \BAE{-R/I}                                                          & 31.60 \small{(0.820)}                      & 43.1 \small{(0.946)}                       & 20.4 \small{(0.954)}                       \\
                                      & \BAE{-R+I}                                                          & \textbf{25.57 \small{(0.766)}}             & \textbf{29.0 \small{(0.929)}}              & \textbf{11.8 \small{(0.874)}}              \\ \midrule
    \multirow{6}{*}{\textbf{wordCNN}}  & Original                                                         & 89.06                             & 91.3                              & 93.2                              \\
                                      & \small{TextFooler}                                                       & 48.77 \small{(0.733)}                      & 58.9 \small{(0.889)}                       & 47.6 \small{(0.812)}                       \\
                                      & \BAE{-R}                                                            & 44.43 \small{(0.735)}                      & 51.0 \small{(0.899)}                       & 29.6 \small{(0.843)}                       \\
                                      & \BAE{-I}                                                            & 44.43 \small{(0.876)}                      & 49.8 \small{(0.958)}                       & 15.4 \small{(0.953)}                       \\
                                      & \BAE{-R/I}                                                          & 32.17 \small{(0.818)}                      & 41.5 \small{(0.940)}                       & 13.0 \small{(0.936)}                       \\
                                      & \BAE{-R+I}                                                          & \textbf{27.83 \small{(0.764)}}             & \textbf{31.1 \small{(0.922)}}              & \textbf{8.4 \small{(0.858)}}               \\ \midrule
    \multirow{6}{*}{\textbf{BERT}}     & Original                                                         & 90.66                             & 97.0                              & 97.6                              \\
                                      & \small{TextFooler}                                                       & 36.23 \small{(0.761)}                      & 69.5 \small{(0.858)}                       & 42.8 \small{(0.866)}                       \\
                                      & \BAE{-R}                                                            & 43.87 \small{(0.764)}                      & 77.2 \small{(0.828)}                       & 37.2 \small{(0.824)}                       \\
                                      & \BAE{-I}                                                            & 33.49 \small{(0.862)}                      & 74.6 \small{(0.918)}                       & 32.2 \small{(0.931)}                       \\
                                      & \BAE{-R/I}                                                          & 24.53 \small{(0.826)}                      & 64.0 \small{(0.903)}                       & 23.6 \small{(0.908)}                       \\
                                      & \BAE{-R+I}                                                          & \textbf{24.34 \small{(0.766)}}             & \textbf{58.5 \small{(0.875)}}              & \textbf{20.2 \small{(0.825)}}              \\ \bottomrule
    \end{tabular}
    }
}{%
  \caption{Automatic evaluation of adversarial attacks on MPQA, Subj and TREC datasets. Other details follow those from Table~\ref{tab:results1}. All 4 modes of \BAE{} attacks almost always outperform TextFooler.} \label{tab:results2}
}
\centering
    \ffigbox[0.78\linewidth]
    {
    \begin{subfigure}[t]{0.33\textwidth}
        \raisebox{-2cm}{\includegraphics[width=\textwidth]{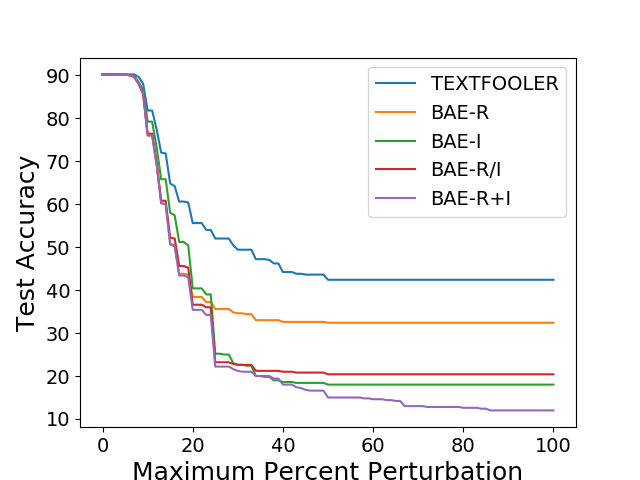}}
        \caption{Word-LSTM}
    \end{subfigure}
    \begin{subfigure}[t]{0.33\textwidth}
        \raisebox{-2cm}{\includegraphics[width=\textwidth]{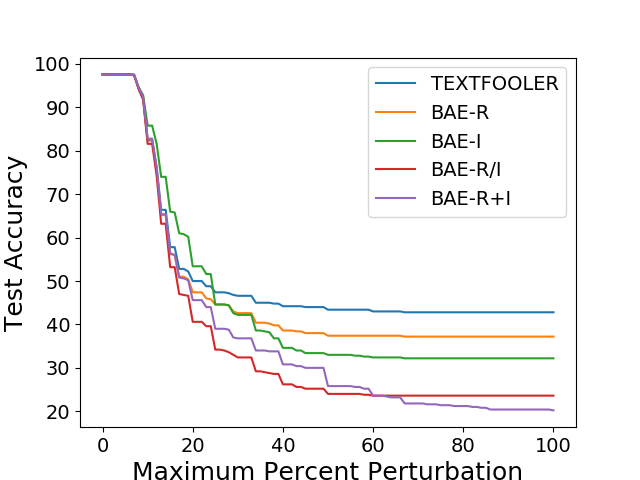}}
        \caption{BERT}
    \end{subfigure}
    
    }
    {
  \caption{Graphs comparing attack effectiveness on the TREC dataset, as a function of maximum \% perturbation to the input.}%
  \label{fig:graphs}
}
\end{floatrow}
\end{figure*}

\myparagraph{Automatic Evaluation Results}
\label{ssec:auto_eval}
We perform the 4 \BAE{} attacks and summarize the results in Tables \ref{tab:results1} and  \ref{tab:results2}. 
Across datasets and models, our \BAE{} attacks are almost always more effective than the baseline attack, achieving significant drops of 40-80\% in test accuracies, with higher average semantic similarities as shown in parentheses. 

With just one exception, \BAE{-R+I} is the strongest attack since it allows both replacement and insertion at the same token position. 
We observe a general trend that the \BAE{-R} and \BAE{-I} attacks often perform comparably, while the \BAE{-R/I} and \BAE{-R+I} attacks are much stronger.
We observe that the BERT classifier is  more robust to \BAE{} and TextFooler attacks than the word-LSTM and word-CNN possibly due to its large size and pre-training on a large corpus.

The TextFooler attack is sometimes stronger than the \BAE{-R} attack for the BERT classifier. We attribute this to the shared parameter space between the BERT-MLM and the BERT classifier before fine-tuning. The predicted tokens from BERT-MLM may not be able to drastically change the internal representations learned by the BERT classifier, hindering their ability to adversarially affect the classifier prediction. 

Additionally, we make some interesting observations pertaining to the average semantic similarity of the adversarial examples with the original sentences (computed using USE). 
From Tables~\ref{tab:results1}, \ref{tab:results2} we observe that across different models and datasets, all \BAE{} attacks have higher average semantic similarity than TextFooler. 
Notably, the \BAE{-I} attack achieves the highest semantic similarity among all the 4 modes. 
This can be explained by the fact that all tokens of the original sentence are retained, in the original order, in the adversarial example generated by \BAE{-I}. 
Interestingly, we observe that the average semantic similarity of the \BAE{-R+I} attack is always higher than the \BAE{-R} attack. 
This lends support to the importance of the `Insert' operation in ameliorating the effect of the `Replace' operation. We further investigate this through an ablation study discussed later.

\begin{table*}
\RawFloats
\parbox{.63\linewidth}{
\centering

\resizebox{\linewidth}{!}{
\begin{tabular}{l}
\toprule
\textbf{Original [Positive Sentiment]:} This film offers many delights and surprises. \\
\textbf{\small{TextFooler}:} This \blue{flick} \blue{citations} \blue{disparate} \blue{revel} and surprises. \\
\textbf{\BAE{-R}:} This \blue{movie} offers \blue{enough} delights and surprises \\
\textbf{\BAE{-I}:} This \red{lovely} film \red{platform} offers many \red{pleasant} delights and surprises \\
\textbf{\BAE{-R/I}:} This \red{lovely} film \blue{serves} \blue{several} \blue{pleasure} and surprises . \\
\textbf{\BAE{-R+I}:} This \red{beautiful} \blue{movie} offers many \red{pleasant} delights and surprises .
\\ \midrule
\textbf{Original [Positive Sentiment]:} Our server was great and we had perfect service.\\
\textbf{\small{TextFooler}:} Our server was \blue{tremendous} and we \blue{assumed} \blue{faultless} \blue{services}. \\
\textbf{\BAE{-R}:} Our server was \blue{decent} and we had \blue{outstanding} service. \\ 
\textbf{\BAE{-I}:}  Our server was great \red{enough} and we had perfect service \red{but}.\\
\textbf{\BAE{-R/I}:}  Our server was great \red{enough} and we \blue{needed} perfect service \red{but}.\\
\textbf{\BAE{-R+I}:} Our server was \blue{decent} \red{company} and we had \blue{adequate} service. \\
\bottomrule
\end{tabular}
}
\vspace{-0.3em}
\caption{Qualitative examples of each attack on the BERT classifier \\(Replacements: \red{Red}, Inserts: \blue{Blue})
}
\label{tab:examples}

}
\hfill
\parbox{.37\linewidth}{
\centering

\resizebox{\linewidth}{!}{
\begin{tabular}{ccccc}
\toprule
\multirow{2}{*}{\textbf{Dataset}} & \multicolumn{4}{c}{\textbf{Sentiment Accuracy (\%)}}  \\
& Original & TF &  R & R+I  \\
\cmidrule(lr){1-5} 
Amazon                   &       95.7             &   79.1          &   \textbf{85.2}         &       83.8  \\ 
IMDB                     &        90.3            &         83.1         &      \textbf{84.3}  &               79.3    \\
MR    &        93.3           &         82.0         &     \textbf{84.6}   &      82.4 \\
\midrule 
\multirow{2}{*}{\textbf{Dataset}} & \multicolumn{4}{c}{\textbf{Naturalness (1-5)}}  \\
& Original & TF &  R & R+I  \\
\cmidrule(lr){1-5} 
Amazon              &       4.26             &   3.17         &   \textbf{3.91}  & 3.71  \\ 
IMDB                    &       4.35           &         3.41        &      \textbf{3.89}  &               3.76   \\
MR     &        4.19         &         3.35         &     \textbf{3.84}   &     3.74    \\ 
\bottomrule
\end{tabular}
}
\vspace{-0.3em}
\caption{Human evaluation results (TF: TextFooler and R(R+I): \BAE{-R(R+I)}).  
}
\label{tab:human}

}
\end{table*}

\myparagraph{Effectiveness} 
We study the effectiveness of \BAE{} on limiting the number of R/I operations permitted on the original text. 
We plot the attack performance as a function of maximum $\%$ perturbation (ratio of number of word replacements and insertions to the length of the original text) for the TREC dataset. 
From Figure~\ref{fig:graphs}, we clearly observe that the \BAE{} attacks are consistently stronger than TextFooler.
The classifier models are relatively robust to perturbations up to 20$\%$, while the effectiveness saturates at 40-50$\%$.
Surprisingly, a 50$\%$ perturbation for the TREC dataset translates to replacing or inserting just 3-4 words, due to the short text lengths. 

\myparagraph{Qualitative Examples}
We present adversarial examples generated by the attacks on sentences from the IMDB and Yelp datasets in Table~\ref{tab:examples}. 
All attack strategies successfully changed the classification to negative, however the \BAE{} attacks produce more natural looking examples than TextFooler.
The tokens predicted by the BERT-MLM fit well in the sentence context, while TextFooler tends to replace words with complex synonyms, which can be easily detected.  
Moreover, \BAE{}'s additional degree of freedom to insert tokens allows for a successful attack with fewer perturbations.

\myparagraph{Human Evaluation}
We perform human evaluation of our \BAE{} attacks on the BERT classifier. For 3 datasets, we consider 100 samples from each test set shuffled randomly with their successful adversarial examples from \BAE{-R}, \BAE{-R+I} and TextFooler. 
We calculate the sentiment accuracy by asking 3 annotators to predict the sentiment for each sentence in this shuffled set.
To evaluate the \textit{naturalness} of the adversarial examples,
we first present the annotators with 50 other original data samples to get a sense of the data distribution. 
We then ask them to score each sentence (on a Likert scale of 1-5) in the shuffled set on its grammar and likelihood of being from the original data. 
We average the 3 scores and present them in Table~\ref{tab:human}.

\begin{table}[b!]
\centering
\resizebox{\textwidth}{!}{
\begin{tabular}{cccccccccc}
\toprule
\multirow{2}{*}{\textbf{Dataset}} & \multicolumn{3}{c}{\textbf{Word-LSTM}} & \multicolumn{3}{c}{\textbf{Word-CNN}} & \multicolumn{3}{c}{\textbf{BERT}} \\ 
\cmidrule(lr){2-4} \cmidrule(lr){5-7} \cmidrule(lr){8-10}
                                  & \textbf{\A}         & \textbf{\B}        &
               \textbf{\C}        & 
               \textbf{\A}        & \textbf{\B}        &
               \textbf{\C}        &
               \textbf{\A}      & \textbf{\B}
               &
               \textbf{\C}        \\ 
               \midrule
\textbf{MR}                       & 15.1               &  10.1       &   3.1     & 
12.4 & 9.6       & 2.8       &       24.3          &  12.9  & 5.7               \\
\textbf{Subj}                     & 14.4                & 12.3        &    5.1    & 16.2               & 13.8         &  7.4     &    13.9             &   11.4   &     7.5        \\ 
\textbf{TREC}                     & 16.6                & 1.6        &    0.2    & 20.0               & 5.0         &  1.4     &    14.0             &   8.6   &     2.4        \\ \bottomrule
\end{tabular}
}
\vspace{-0.75em}
\caption{
Analyzing relative importance of `Replace' and `Insert' perturbations for \BAE. 
\A \ denotes \% of test instances which are successfully attacked by \BAE{-R/I}, but not \BAE{-R}, i.e. \A \ : $(\BAEfont{R/I})\cap\overline{\BAEfont{R}}$. 
Similarly, \
 \B \ : $(\BAEfont{R/I})\cap\overline{\BAEfont{I}}$  \ and \
\C \ : $(\BAEfont{R/I})\cap\overline{\BAEfont{R}}\cap\overline{\BAEfont{I}}$. 
}
\label{tab:ablation}
\end{table}

Both \BAE{-R} and \BAE{-R+I} attacks almost always outperform TextFooler in both metrics. 
\BAE{-R} outperforms \BAE{-R+I} since the latter inserts tokens to strengthen the attack, at the expense of naturalness and sentiment accuracy. 
Interestingly, the \BAE{-R+I} attacks achieve higher average semantic similarity scores than \BAE{-R}, as discussed in Section \ref{ssec:auto_eval}.
This exposes the shortcomings of using USE for evaluating the retention of semantics of adversarial examples, and reiterates the importance of human-centered evaluation. 
The gap between the scores on the original data and the adversarial examples speaks for the limitations of the attacks, however \BAE{} represents an important step forward towards improved adversarial examples.

\myparagraph{Replace vs. Insert} 
\label{subsec:ablation}
Our \BAE{} attacks allow insertion operations in addition to replace. 
We analyze the benefits of this flexibility of R/I operations in Table~\ref{tab:ablation}.
From Table~\ref{tab:ablation}, the splits \A \ and \B \ are the $\%$ of test points which \emph{compulsorily} need I and R operations respectively for a successful attack.
We can observe that the split \A \ is larger than \B \ thereby indicating the importance of the I operation over R.
Test points in split \C \ require both R \emph{and} I operations for a successful attack. 
Interestingly, split \C \ is largest for Subj, which is the most robust to attack (Table~\ref{tab:results2}) and hence needs both R/I operations. Thus, this study gives positive insights towards the importance of having the flexibility to both replace and insert words.

We present complete effectiveness graphs and details of human evaluation in Appendix~\ref{app:full_results} and \ref{app:human_eval}. \BAE{} is implemented\footnote{\url{https://github.com/QData/TextAttack/blob/master/textattack/attack_recipes/bae_garg_2019.py}} in TextAttack~\cite{morris2020textattack}, a popular suite of NLP adversarial attacks. 

\section{Conclusion}
\label{sec:conclusion}
In this paper, we have presented a new technique for generating adversarial examples (\BAE) through contextual perturbations based on the BERT Masked Language Model. 
We propose inserting and/or replacing tokens from a sentence, in their order of importance for the text classification task, using a BERT-MLM.
Automatic and human evaluation on several datasets demonstrates the strength and effectiveness of our attack.

\section*{Acknowledgments}

The authors thank Arka Sadhu, Kalpesh Krishna, Aws Albarghouthi, Yingyu Liang and Justin Hsu for providing in-depth
feedback for this research. The authors thank Jack Morris and Jin Yong Yoo for integrating \BAE{} in the TextAttack framework.
This work is supported, in part, by the National Science Foundation CCF under award 1652140.

\section*{Broader Ethical Impact}
Our work addresses the important problem of adversarial vulnerabilities of modern text classification models. 
While we acknowledge the possibility of its misuse to maliciously attack publicly available text classifiers, we believe our work represents an important step forward in analyzing the robustness of NLP models.
We hope our work inspires improved defenses against adversarial attacks on text classification models.

\bibliographystyle{acl_natbib}
\bibliography{emnlp2020}

\begin{thebibliography}{29}
\expandafter\ifx\csname natexlab\endcsname\relax\def\natexlab#1{#1}\fi

\bibitem[{Alzantot et~al.(2018)Alzantot, Sharma, Elgohary, Ho, Srivastava, and
  Chang}]{alzantot-etal-2018-generating}
Moustafa Alzantot, Yash Sharma, Ahmed Elgohary, Bo-Jhang Ho, Mani Srivastava,
  and Kai-Wei Chang. 2018.
\newblock \href {https://doi.org/10.18653/v1/D18-1316} {Generating natural
  language adversarial examples}.
\newblock In \emph{Proceedings of the 2018 Conference on Empirical Methods in
  Natural Language Processing}. Association for Computational Linguistics.

\bibitem[{Cer et~al.(2018)Cer, Yang, Kong, Hua, Limtiaco, John, Constant,
  Guajardo{-}Cespedes, Yuan, Tar, Sung, Strope, and
  Kurzweil}]{DBLP:journals/corr/abs-1803-11175}
Daniel Cer, Yinfei Yang, Sheng{-}yi Kong, Nan Hua, Nicole Limtiaco, Rhomni~St.
  John, Noah Constant, Mario Guajardo{-}Cespedes, Steve Yuan, Chris Tar,
  Yun{-}Hsuan Sung, Brian Strope, and Ray Kurzweil. 2018.
\newblock \href {http://arxiv.org/abs/1803.11175} {Universal sentence encoder}.

\bibitem[{Devlin et~al.(2018)Devlin, Chang, Lee, and
  Toutanova}]{devlin2018bert}
Jacob Devlin, Ming-Wei Chang, Kenton Lee, and Kristina Toutanova. 2018.
\newblock Bert: Pre-training of deep bidirectional transformers for language
  understanding.
\newblock \emph{arXiv preprint arXiv:1810.04805}.

\bibitem[{Ebrahimi et~al.(2018)Ebrahimi, Rao, Lowd, and
  Dou}]{ebrahimi-etal-2018-hotflip}
Javid Ebrahimi, Anyi Rao, Daniel Lowd, and Dejing Dou. 2018.
\newblock \href {https://doi.org/10.18653/v1/P18-2006} {{H}ot{F}lip: White-box
  adversarial examples for text classification}.
\newblock In \emph{Proceedings of the 56th Annual Meeting of the Association
  for Computational Linguistics}, pages 31--36, Melbourne, Australia.
  Association for Computational Linguistics.

\bibitem[{Feng et~al.(2018)Feng, Wallace, Iyyer, Rodriguez, II, and
  Boyd{-}Graber}]{DBLP:journals/corr/abs-1804-07781}
Shi Feng, Eric Wallace, Mohit Iyyer, Pedro Rodriguez, Alvin~Grissom II, and
  Jordan~L. Boyd{-}Graber. 2018.
\newblock \href {http://arxiv.org/abs/1804.07781} {Right answer for the wrong
  reason: Discovery and mitigation}.
\newblock \emph{CoRR}, abs/1804.07781.

\bibitem[{Gao et~al.(2018)Gao, Lanchantin, Soffa, and
  Qi}]{DBLP:journals/corr/abs-1801-04354}
Ji~Gao, Jack Lanchantin, Mary~Lou Soffa, and Yanjun Qi. 2018.
\newblock \href {http://arxiv.org/abs/1801.04354} {Black-box generation of
  adversarial text sequences to evade deep learning classifiers}.
\newblock \emph{CoRR}, abs/1801.04354.

\bibitem[{Goodfellow et~al.(2015)Goodfellow, Shlens, and Szegedy}]{43405}
Ian Goodfellow, Jonathon Shlens, and Christian Szegedy. 2015.
\newblock \href {http://arxiv.org/abs/1412.6572} {Explaining and harnessing
  adversarial examples}.
\newblock In \emph{International Conference on Learning Representations}.

\bibitem[{Hochreiter and
  Schmidhuber(1997)}]{Hochreiter:1997:LSM:1246443.1246450}
Sepp Hochreiter and J\"{u}rgen Schmidhuber. 1997.
\newblock \href {https://doi.org/10.1162/neco.1997.9.8.1735} {Long short-term
  memory}.
\newblock \emph{Neural Com.}, 9(8):1735--1780.

\bibitem[{Jin et~al.(2019)Jin, Jin, Zhou, and Szolovits}]{jin2019bert}
Di~Jin, Zhijing Jin, Joey~Tianyi Zhou, and Peter Szolovits. 2019.
\newblock Is bert really robust? natural language attack on text classification
  and entailment.
\newblock \emph{arXiv preprint arXiv:1907.11932}.

\bibitem[{Kim(2014)}]{kim2014convolutional}
Yoon Kim. 2014.
\newblock \href {http://aclweb.org/anthology/D/D14/D14-1181.pdf} {Convolutional
  neural networks for sentence classification}.
\newblock In \emph{Proceedings of the 2014 Conference on Empirical Methods in
  Natural Language Processing}, pages 1746--1751.

\bibitem[{Kurakin et~al.(2017)Kurakin, Goodfellow, and Bengio}]{45818}
Alexey Kurakin, Ian Goodfellow, and Samy Bengio. 2017.
\newblock \href {https://arxiv.org/abs/1607.02533} {Adversarial examples in the
  physical world}.
\newblock \emph{ICLR Workshop}.

\bibitem[{Li et~al.(2020)Li, Zhang, Peng, Chen, Brockett, Sun, and
  Dolan}]{li2020contextualized}
Dianqi Li, Yizhe Zhang, Hao Peng, Liqun Chen, Chris Brockett, Ming-Ting Sun,
  and Bill Dolan. 2020.
\newblock \href {http://arxiv.org/abs/2009.07502} {Contextualized perturbation
  for textual adversarial attack}.

\bibitem[{Li et~al.(2016)Li, Monroe, and Jurafsky}]{DBLP:journals/corr/LiMJ16a}
Jiwei Li, Will Monroe, and Dan Jurafsky. 2016.
\newblock \href {http://arxiv.org/abs/1612.08220} {Understanding neural
  networks through representation erasure}.
\newblock \emph{CoRR}, abs/1612.08220.

\bibitem[{Li and Roth(2002)}]{Li:2002:LQC:1072228.1072378}
Xin Li and Dan Roth. 2002.
\newblock \href {https://doi.org/10.3115/1072228.1072378} {Learning question
  classifiers}.
\newblock In \emph{Proceedings of the 19th International Conference on
  Computational Linguistics - Volume 1}, COLING '02, pages 1--7, Stroudsburg,
  PA, USA. Association for Computational Linguistics.

\bibitem[{Liang et~al.(2017)Liang, Li, Su, Bian, Li, and
  Shi}]{DBLP:journals/corr/LiangLSBLS17}
Bin Liang, Hongcheng Li, Miaoqiang Su, Pan Bian, Xirong Li, and Wenchang Shi.
  2017.
\newblock \href {http://arxiv.org/abs/1704.08006} {Deep text classification can
  be fooled}.
\newblock \emph{CoRR}, abs/1704.08006.

\bibitem[{Morris et~al.(2020)Morris, Lifland, Yoo, Grigsby, Jin, and
  Qi}]{morris2020textattack}
John~X. Morris, Eli Lifland, Jin~Yong Yoo, Jake Grigsby, Di~Jin, and Yanjun Qi.
  2020.
\newblock \href {http://arxiv.org/abs/2005.05909} {Textattack: A framework for
  adversarial attacks, data augmentation, and adversarial training in nlp}.

\bibitem[{Mrk{\v{s}}i{\'c} et~al.(2017)Mrk{\v{s}}i{\'c}, Vuli{\'c},
  S{\'e}aghdha, Leviant, Reichart, Ga{\v{s}}i{\'c}, Korhonen, and
  Young}]{mrkvsic2017semantic}
Nikola Mrk{\v{s}}i{\'c}, Ivan Vuli{\'c}, Diarmuid~{\'O} S{\'e}aghdha, Ira
  Leviant, Roi Reichart, Milica Ga{\v{s}}i{\'c}, Anna Korhonen, and Steve
  Young. 2017.
\newblock Semantic specialization of distributional word vector spaces using
  monolingual and cross-lingual constraints.
\newblock \emph{Transactions of the Association for Computational Linguistics}.

\bibitem[{Mrk\v{s}i\'c et~al.(2016)Mrk\v{s}i\'c, {\'O S\'eaghdha}, Thomson,
  Ga\v{s}i\'c, Rojas-Barahona, Su, Vandyke, Wen, and Young}]{mrksic:2016:naacl}
Nikola Mrk\v{s}i\'c, Diarmuid {\'O S\'eaghdha}, Blaise Thomson, Milica
  Ga\v{s}i\'c, Lina Rojas-Barahona, Pei-Hao Su, David Vandyke, Tsung-Hsien Wen,
  and Steve Young. 2016.
\newblock Counter-fitting word vectors to linguistic constraints.
\newblock In \emph{Proceedings of HLT-NAACL}.

\bibitem[{Pang and Lee(2004)}]{Pang+Lee:04a}
Bo~Pang and Lillian Lee. 2004.
\newblock A sentimental education: Sentiment analysis using subjectivity.
\newblock In \emph{Proceedings of ACL}, pages 271--278.

\bibitem[{Pang and Lee(2005)}]{Pang+Lee:05a}
Bo~Pang and Lillian Lee. 2005.
\newblock Seeing stars: Exploiting class relationships for sentiment
  categorization with respect to rating scales.
\newblock In \emph{Proceedings of ACL}.

\bibitem[{Papernot et~al.(2017)Papernot, McDaniel, Goodfellow, Jha, Celik, and
  Swami}]{Papernot:2017:PBA:3052973.3053009}
Nicolas Papernot, Patrick McDaniel, Ian Goodfellow, Somesh Jha, Z.~Berkay
  Celik, and Ananthram Swami. 2017.
\newblock \href {https://doi.org/10.1145/3052973.3053009} {Practical black-box
  attacks against machine learning}.
\newblock In \emph{Proceedings of the 2017 ACM on Asia Conference on Computer
  and Communications Security}, ASIA CCS '17, pages 506--519, New York, NY,
  USA. ACM.

\bibitem[{Radford et~al.(2019)Radford, Wu, Child, Luan, Amodei, and
  Sutskever}]{radford2019language}
Alec Radford, Jeff Wu, Rewon Child, David Luan, Dario Amodei, and Ilya
  Sutskever. 2019.
\newblock Language models are unsupervised multitask learners.

\bibitem[{Ren et~al.(2019)Ren, Deng, He, and Che}]{ren-etal-2019-generating}
Shuhuai Ren, Yihe Deng, Kun He, and Wanxiang Che. 2019.
\newblock \href {https://doi.org/10.18653/v1/P19-1103} {Generating natural
  language adversarial examples through probability weighted word saliency}.
\newblock In \emph{Proceedings of the 57th Annual Meeting of the Association
  for Computational Linguistics}. Association for Computational Linguistics.

\bibitem[{Sarma et~al.(2018)Sarma, Liang, and Sethares}]{sarma2018domain}
Prathusha~K Sarma, Yingyu Liang, and Bill Sethares. 2018.
\newblock Domain adapted word embeddings for improved sentiment classification.
\newblock In \emph{Proceedings of the 56th Annual Meeting of the Association
  for Computational Linguistics}, pages 37--42.

\bibitem[{Shi and Huang(2020)}]{shi2020robustness}
Zhouxing Shi and Minlie Huang. 2020.
\newblock \href {http://arxiv.org/abs/1909.02560} {Robustness to modification
  with shared words in paraphrase identification}.

\bibitem[{Szegedy et~al.(2014)Szegedy, Zaremba, Sutskever, Bruna, Erhan,
  Goodfellow, and Fergus}]{42503}
Christian Szegedy, Wojciech Zaremba, Ilya Sutskever, Joan Bruna, Dumitru Erhan,
  Ian Goodfellow, and Rob Fergus. 2014.
\newblock \href {http://arxiv.org/abs/1312.6199} {Intriguing properties of
  neural networks}.
\newblock In \emph{International Conference on Learning Representations}.

\bibitem[{Wiebe and Wilson(2005)}]{Wiebe2005}
Janyce Wiebe and Theresa Wilson. 2005.
\newblock \href {https://doi.org/10.1007/s10579-005-7880-9} {Annotating
  expressions of opinions and emotions in language}.
\newblock \emph{Language Resources and Evaluation}.

\bibitem[{Zhang et~al.(2019)Zhang, Sheng, and
  Alhazmi}]{DBLP:journals/corr/abs-1901-06796}
Wei~Emma Zhang, Quan~Z. Sheng, and Ahoud Abdulrahmn~F. Alhazmi. 2019.
\newblock \href {http://arxiv.org/abs/1901.06796} {Generating textual
  adversarial examples for deep learning models: {A} survey}.
\newblock \emph{CoRR}, abs/1901.06796.

\bibitem[{Zhao et~al.(2018)Zhao, Dua, and Singh}]{zhao2018generating}
Zhengli Zhao, Dheeru Dua, and Sameer Singh. 2018.
\newblock \href {https://openreview.net/forum?id=H1BLjgZCb} {Generating natural
  adversarial examples}.
\newblock In \emph{International Conference on Learning Representations}.

\end{thebibliography}

\appendix

\textbf{\Large Appendix}

\section{Experimental Reproducibility}
\label{app:exp_details}

\paragraph{Dataset and Models}
The dataset statistics are reported in Table~\ref{tab:dataset} and we give a brief overview of the dataset and the task for which it is used along with public links to download the datasets.
\begin{itemize}
    \item \textit{Amazon}: Amazon product reviews dataset
    \footnote{\url{https://archive.ics.uci.edu/ml/datasets/Sentiment+Labelled+Sentences}}. 
    \item \textit{Yelp}: A restaurant reviews dataset from Yelp\footnotemark[2]. 
    \item \textit{IMDB}: IMDB movie reviews dataset\footnotemark[2].
    \item \textit{MR}: A movie reviews dataset based on subjective rating and sentiment polarity  \footnote{\url{https://www.cs.cornell.edu/people/pabo/movie-review-data/}}.
    \item \textit{MPQA}: An unbalanced dataset for polarity detection of opinions \footnote{\url{http://mpqa.cs.pitt.edu/}}.
    \item \textit{TREC}: A dataset for classifying types of questions with 6 classes \footnote{\url{http://cogcomp.org/Data/QA/QC/}}. 
    \item \textit{SUBJ}: A dataset for classifying a sentence as objective or subjective.
    \footnotemark[2]
\end{itemize}

\begin{table}[t]
\renewcommand\thetable{5}
\centering
\resizebox{\linewidth}{!}{
\begin{tabular}{ccccc}
\toprule
\textbf{Dataset} & \textbf{\# Classes} & \textbf{Train} & \textbf{Test} & \textbf{Avg Length} \\ \midrule
Amazon           & 2          & 900            & 100           &         10.29            \\ 
Yelp             & 2          & 900            & 100           &         11.66           \\ 
IMDB             & 2          & 900            & 100           &         17.56            \\ 
MR               & 2          & 9595           & 1067          &         20.04          \\ 
MPQA             & 2          & 9543           & 1060          &         3.24          \\ 
Subj             & 2          & 9000           & 1000          &         23.46              \\ 
TREC             & 6          & 5951           & 500           &         7.57            \\  \bottomrule
\end{tabular}
}
\caption{Summary statistics for the datasets}
\label{tab:dataset}
\end{table}

\paragraph{Training Details}
On the sentence classification task, we target three models: word-based convolutional neural network (WordCNN), word-based LSTM, and the state-of-the-art BERT. We use 100 filters of sizes 3,4,5 for the WordCNN model with a dropout of 0.3. Similar to \cite{jin2019bert} we use a 1-layer bi-directional LSTM with 150 hidden units and a dropout of 0.3. For both models, we use the 300 dimensional pre-trained counter fitted word embeddings~\cite{mrkvsic2017semantic}. 

For the BERT classifier, we used the BERT base uncased model which has 12-layers, 12 attention heads and 768 hidden dimension size. Across all models and datasets, we use the standard BERT uncased vocabulary of size 30522. We first train all three models on the training data split and use early stopping on the test dataset. For BERT fine-tuning, we use the standard setting of an Adam classifier having a learning rate of $2 \times 10^{-5}$ and 2 fine-tuning epochs.

For our \BAE{} attacks, we use a pre-trained BERT Base-uncased MLM to predict the masked tokens. We only consider the top K=50 synonyms from the BERT-MLM predictions and set a threshold of 0.8 for the cosine similarity between USE based embeddings of the adversarial and input text.

For R operations, we filter out predicted tokens which form a different POS than the original token in the sentence. For both R and I operations, we filter out stop words using NLTK from the set of predicted tokens. Additionally we filter out antonyms using synonym embeddings~\cite{mrksic:2016:naacl} for sentiment analysis tasks. 

\section{Results}
\label{app:full_results}
\begin{figure*}[h!]
    \begin{subfigure}[h]{0.2\textwidth}
        \includegraphics[width=\textwidth]{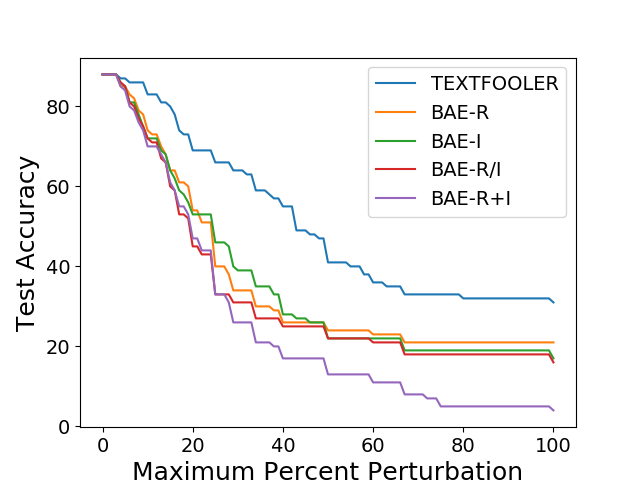}
        \caption{Word-LSTM}
    \end{subfigure}
    \begin{subfigure}[h]{0.2\textwidth}
        \includegraphics[width=\textwidth]{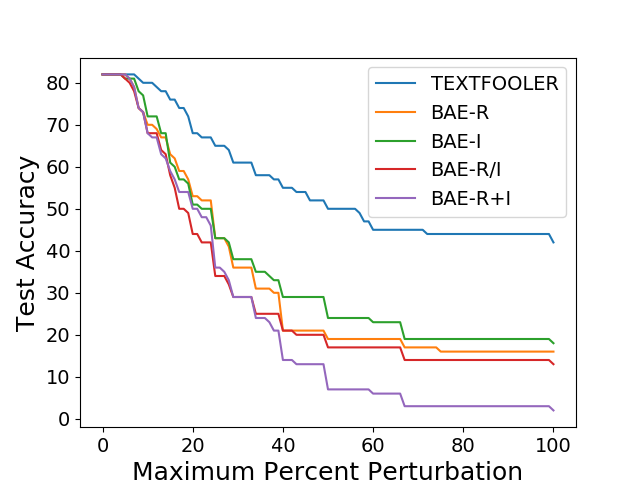}
        \caption{Word-CNN}
    \end{subfigure}
    \begin{subfigure}[h]{0.2\textwidth}
        \includegraphics[width=\textwidth]{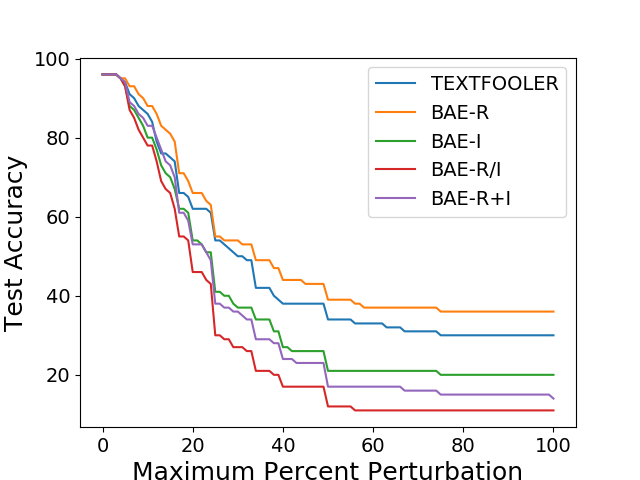}
        \caption{BERT}
    \end{subfigure}
    \vspace{-8pt}
    \caption{Amazon}
    \label{app:graph_amazon}
\end{figure*}

\begin{figure*}[h]
    \begin{subfigure}[h]{0.2\textwidth}
        \includegraphics[width=\textwidth]{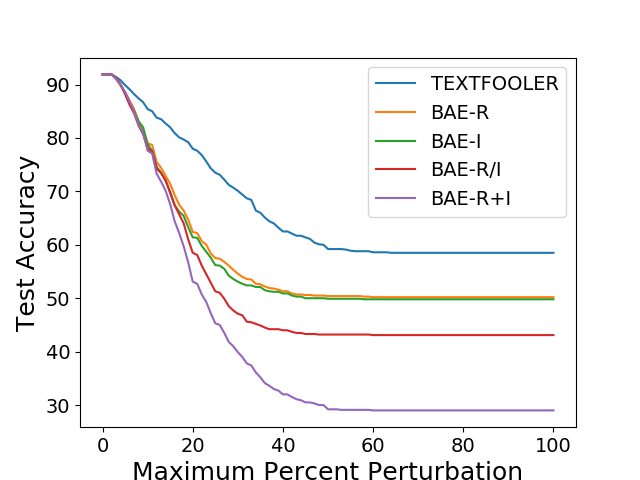}
        \caption{Word-LSTM}
    \end{subfigure}
    \begin{subfigure}[h]{0.2\textwidth}
        \includegraphics[width=\textwidth]{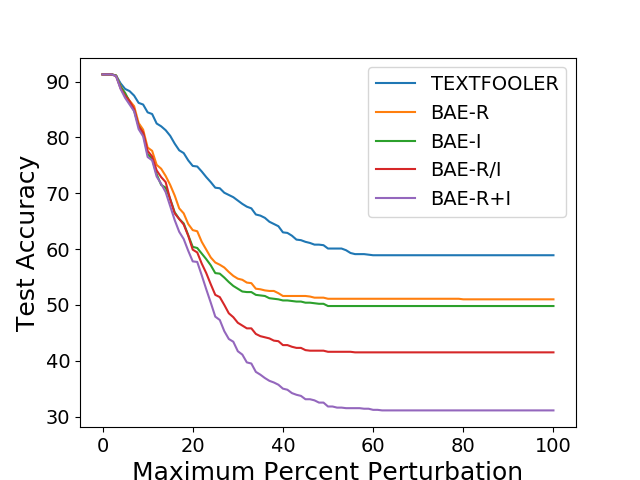}
        \caption{Word-CNN}
    \end{subfigure}
    \begin{subfigure}[h]{0.2\textwidth}
        \includegraphics[width=\textwidth]{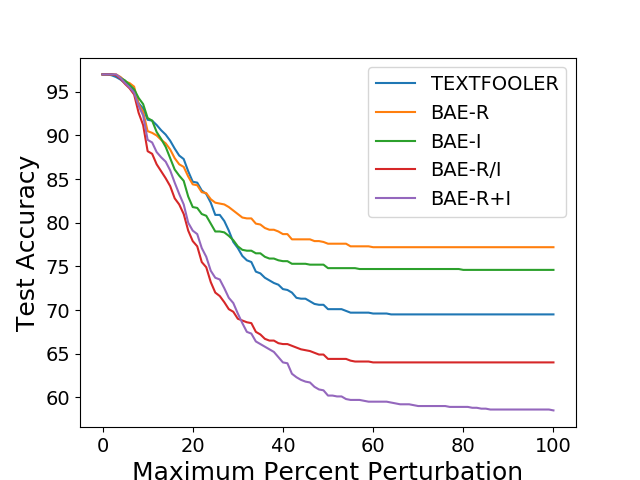}
        \caption{BERT}
    \end{subfigure}
    \vspace{-8pt}
    \caption{Subj}
    \label{app:graph_subj}
\end{figure*}

\vspace{-0.5em}
Figures~\ref{app:graph_amazon} - \ref{app:graph_subj} are the complete set of graphs showing the attack effectiveness for all seven datasets.

\section{Human Evaluation}
\label{app:human_eval}
\vspace{-0.5em}
We ask the human evaluators to judge the \textit{naturalness} of texts presented to them, i.e. whether they think they are adversarial examples or not.
They were instructed to do so on the basis of grammar and how likely they think it is from the original dataset, and rate each example on the following Likert scale of 1-5: 

1) Sure adversarial sample

2) Likely an adversarial example

3) Neutral

4) Likely an original sample

5) Sure original sample. 

From the results of Table 3, it is clear that \BAE{-R} always beats the sentiment accuracy and \textit{naturalness} score of TextFooler. The latter is due to unnaturally long and complex synonym replacements on using TextFooler.


\vfill


\begin{figure*}[h]
    \begin{subfigure}[h]{0.2\textwidth}
        \includegraphics[width=\textwidth]{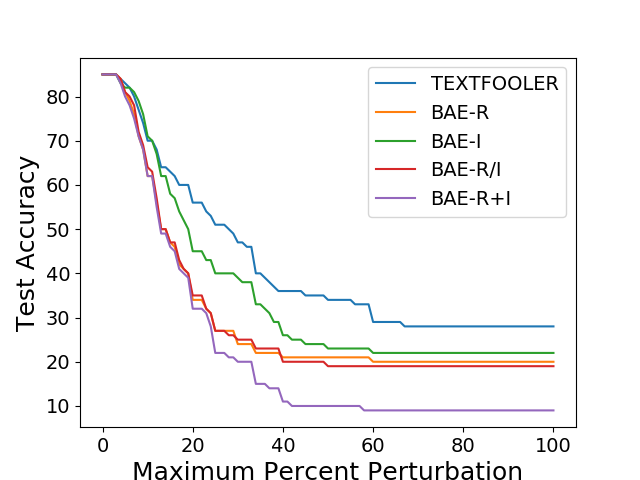}
        \caption{Word-LSTM}
    \end{subfigure}
    \begin{subfigure}[h]{0.2\textwidth}
        \includegraphics[width=\textwidth]{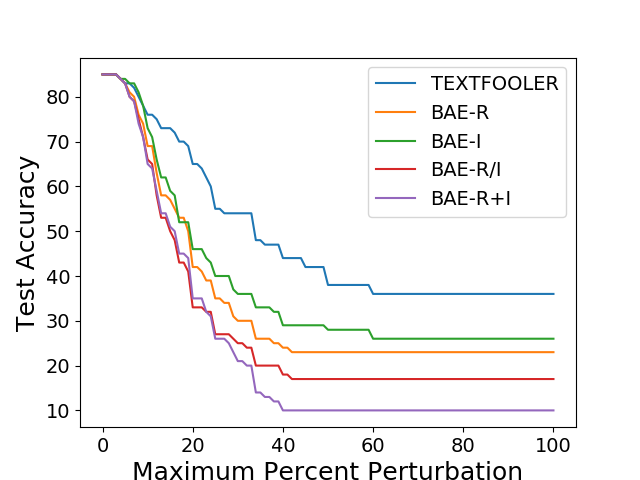}
        \caption{Word-CNN}
    \end{subfigure}
    \begin{subfigure}[h]{0.2\textwidth}
        \includegraphics[width=\textwidth]{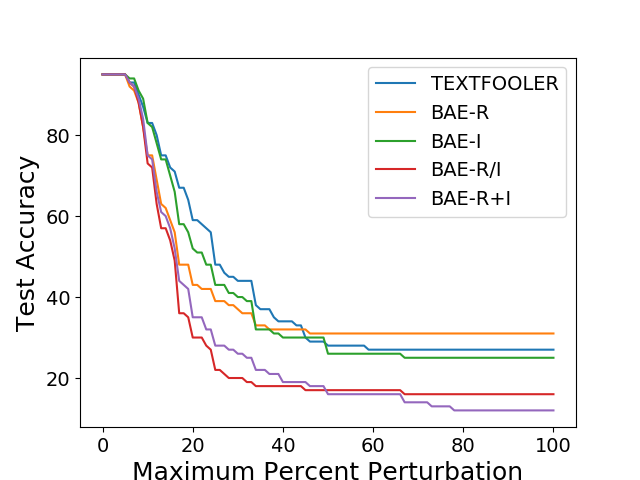}
        \caption{BERT}
    \end{subfigure}
    \vspace{-8pt}
    \caption{Yelp}
\end{figure*}   

\vfill

\begin{figure*}[h]
    \begin{subfigure}[h]{0.2\textwidth}
        \includegraphics[width=\textwidth]{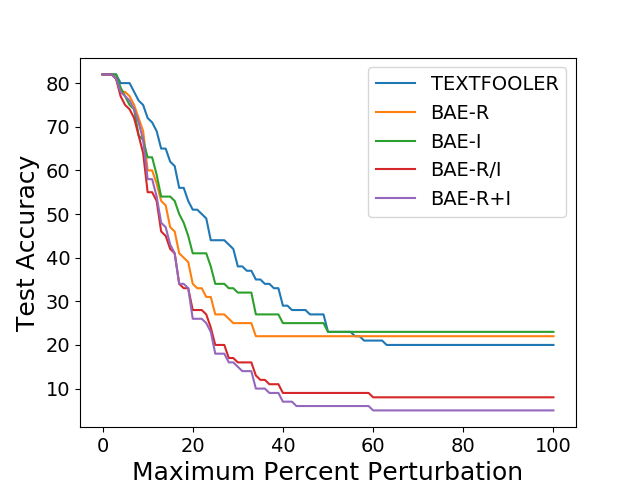}
        \caption{Word-LSTM}
    \end{subfigure}
    \begin{subfigure}[h]{0.2\textwidth}
        \includegraphics[width=\textwidth]{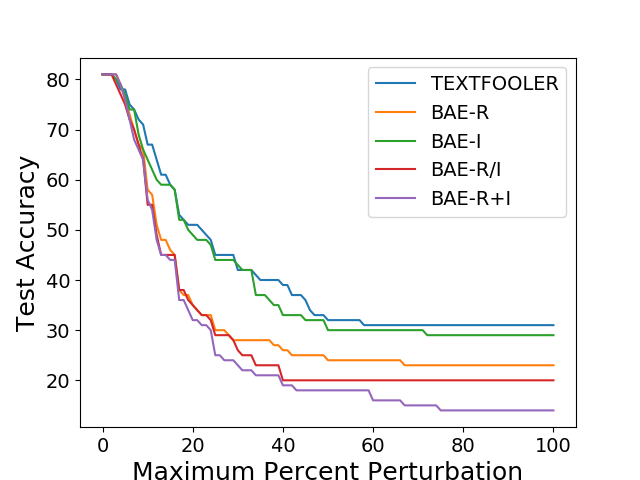}
        \caption{Word-CNN}
    \end{subfigure}
    \begin{subfigure}[h]{0.2\textwidth}
        \includegraphics[width=\textwidth]{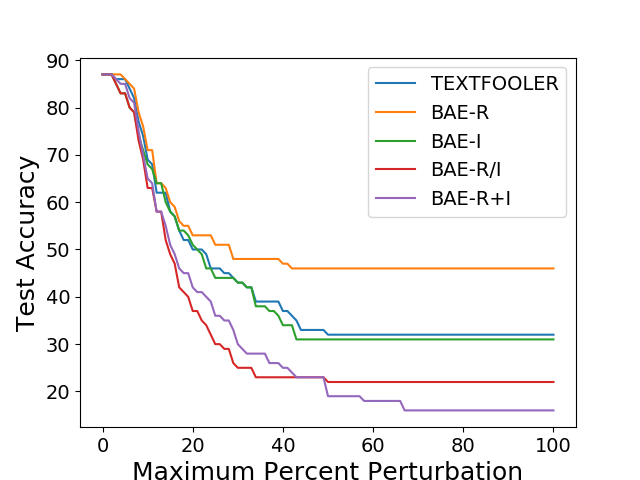}
        \caption{BERT}
    \end{subfigure}
    \vspace{-8pt}
    \caption{IMDB}
\end{figure*}   

\vfill

\begin{figure*}[h]
    \begin{subfigure}[h]{0.2\textwidth}
        \includegraphics[width=\textwidth]{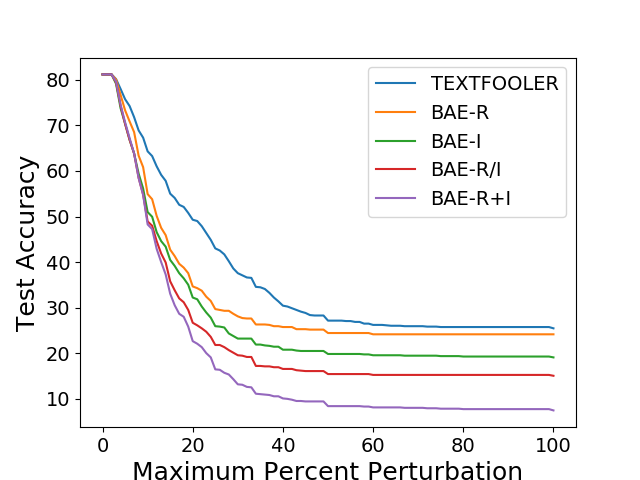}
        \caption{Word-LSTM}
    \end{subfigure}
    \begin{subfigure}[h]{0.2\textwidth}
        \includegraphics[width=\textwidth]{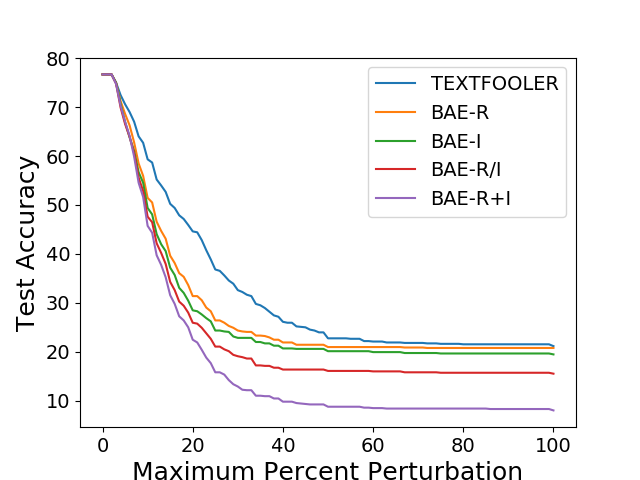}
        \caption{Word-CNN}
    \end{subfigure}
    \begin{subfigure}[h]{0.2\textwidth}
        \includegraphics[width=\textwidth]{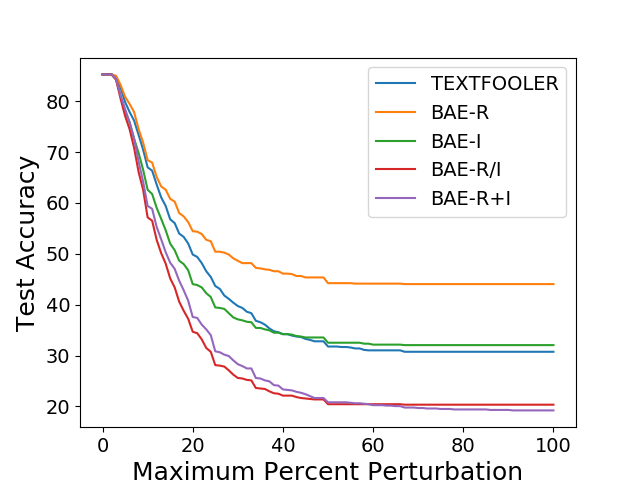}
        \caption{BERT}
    \end{subfigure}
    \vspace{-8pt}
    \caption{MR}
\end{figure*}   

\vfill

\begin{figure*}[h]
    \begin{subfigure}[h]{0.2\textwidth}
        \includegraphics[width=\textwidth]{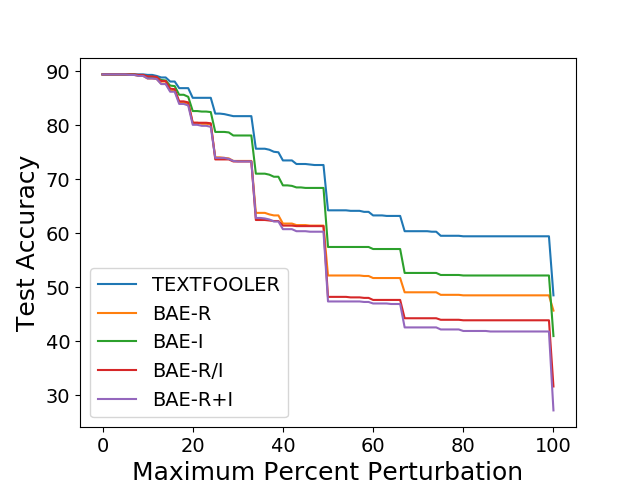}
        \caption{Word-LSTM}
    \end{subfigure}
    \begin{subfigure}[h]{0.2\textwidth}
        \includegraphics[width=\textwidth]{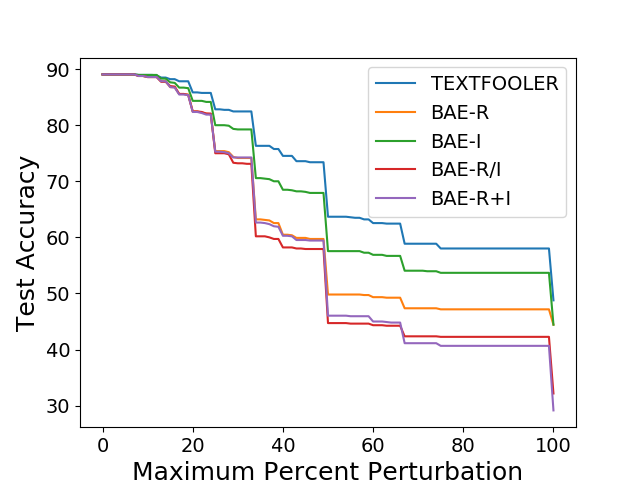}
        \caption{Word-CNN}
    \end{subfigure}
    \begin{subfigure}[h]{0.2\textwidth}
        \includegraphics[width=\textwidth]{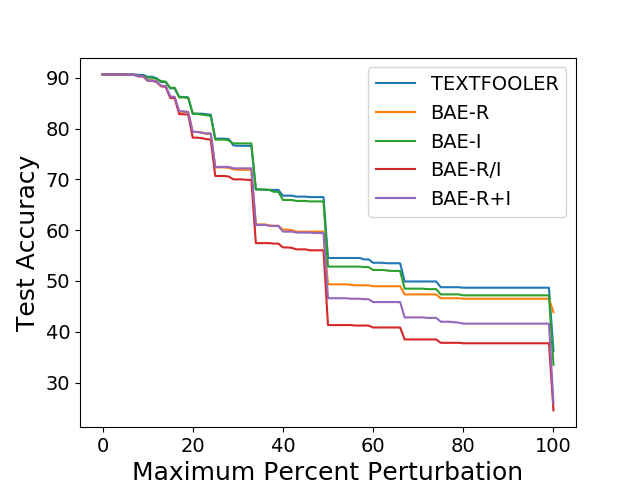}
        \caption{BERT}
    \end{subfigure}
    \vspace{-8pt}
    \caption{MPQA}
\end{figure*}   

\vfill

\end{document}